\documentclass{jhl}
\definecolor{aa}{RGB}{46,111,165}
\usepackage{adjustbox}
\usepackage{multirow}
\usepackage{multicol}
\usepackage{booktabs}
\usepackage{dblfloatfix}
\usepackage{placeins}

\title{RACE-FPP: A Robust AI-assisted Characterisation Enhancement for Fringe Projection Profilometry}

\author[thu,corr]{Osman Ali}
\author[thu]{Xiangjun Kong}
\author[thu]{Tibebe Yalew}
\author[fsc]{Waiel Elmadih}
\author[thu,]{Samanta Piano}

\address[thu]{Manufacturing Metrology Team, Faculty of Engineering, University of Nottingham, Nottingham, NG8 1BB, UK}
\address[fsc]{Taraz Metrology Ltd., Strelley Hall, Nottingham NG8 6PE, UK}

\thanks[corr]{Correspondence to: Osman Ali: \url{osman.ali@nottingham.ac.uk}}

\shortauthor{Ali, et al.}
\shorttitle{preprint}
\doi{}

\begin{abstract}
    Fringe Projection Profilometry (FPP) requires precise system characterisation to achieve reliable three-dimensional (3D) reconstructions; however, characterisation accuracy strongly depends on robust checkerboard feature localisation, which can deteriorate under challenging imaging conditions such as lens blur and characterisation target orientations. Existing deep learning-based corner detectors are typically assessed using detection metrics and camera reprojection error alone, without considering their wider impact on projector characterisation, camera-projector stereo characterisation consistency, or overall measurement accuracy. In this work, we introduce a complete FPP characterisation pipeline that incorporates deep learning-based corner detection into the standard camera characterisation workflow. We also characterise the projector by sampling phase values at the centres of the white squares in the characterisation target. Rather than treating corner detection as an isolated task, the proposed framework explicitly analyses how localisation errors propagate throughout the entire FPP characterisation chain. Performance is evaluated using detection metrics (e.g., precision and recall), camera and projector reprojection errors, and the camera and projector stereo characterisation. Across a mixed dataset of clean and degraded images, the camera reprojection error is reduced from 1.237 pixels to 0.259 pixels, while the projector reprojection error is reduced by roughly 50\%. Dimensional evaluation of reconstructed artefacts shows improved geometric accuracy compared with those resulting from the conventional pipeline. Overall, the findings indicate increased robustness of system-level characterisation under challenging imaging conditions, thereby enabling more reliable industrial FPP measurements.
        \keywords{Fringe Projection Profilometry, Camera characterisation, Projector characterisation, Deep learning, Checkerboard corner detection.}
\end{abstract}

\begin{document}
\maketitle

\section*{Introduction}
\noindent
Fringe Projection Profilometry (FPP) is a non-contact optical measurement technique widely used in manufacturing, aerospace, and biomedical applications \cite{xu2020status, shaheen2021characterisation}. In FPP, a projector emits structured fringe patterns onto an object, and a camera captures the deformed patterns to reconstruct its three-dimensional (3D) geometry. Accurate camera and projector characterisation is essential, as errors in intrinsic (focal lengths, principal points, skewness) and extrinsic parameters (camera and projector poses and their stereo relationship) directly affect final reconstruction accuracy \cite{zhang2021flexible, liu2022high}. 
\par \setlength\parindent{1em} Among other characterisation targets, planar checkerboards are commonly used for camera and projector characterisation because of their simple geometry and well-defined feature locations \cite{sun2008robust}. Characterisation quality critically depends on accurate localisation of these corners; even sub-pixel errors can propagate through the characterisation chain and distort 3D reconstruction \cite{chen2021automatic}. In FPP systems, characterisation images are often affected by lens defocus and environment-induced noise (vibrations, light variations, etc.), which make accurate point localisation more difficult \cite{zhang2017sub, lu2016flexible}.
\par \setlength\parindent{1em} Camera characterisation in FPP systems is commonly performed using established techniques such as Zhang’s characterisation method, which estimates intrinsic and extrinsic parameters from multiple views of a planar checkerboard pattern \cite{zhang2002flexible}. While conventional detection algorithms, such as those implemented in OpenCV, perform well under controlled conditions, their performance degrades in practical FPP scenarios \cite{wu2021highly, zuo2025checkerboard}. These limitations introduce biases in the estimated characterisation parameters and reduce reconstruction accuracy. \cite{zhang2002flexible, rao2018local}.
\par \setlength\parindent{1em} To address these challenges, robust checkerboard corner detection becomes essential. This process typically involves global board identification followed by accurate corner localisation with sub-pixel refinement \cite{duda2018accurate}. Classical corner detection approaches can be classified into three categories: intensity-, contour-, and binary-based methods \cite{wang2022robust}. Intensity-based detection is a widely used approach that identifies corners by computing image derivatives (gradients) to locate points where the intensity varies across multiple directions. Examples of intensity-based methods are the Harris detector \cite{harris1988combined} and its variants \cite{wang2008improved, shang2024image, shang2024research}, which provide robustness to rotation and illumination changes but remain computationally demanding and limited in localisation precision. Enhancements combining FAST and SUSAN detectors \cite{wang2015harris, feng2019feature} improve efficiency; however, performance degrades under defocus blur, noise, and extreme board poses commonly encountered in FPP \cite{yuan2023unsupervised, wang2022robust}.
\par \setlength\parindent{1em} Recent advances in deep learning have reformulated corner detection as a regression or detection task using convolutional neural networks (CNNs) \cite{jiang2022photohelper}. Some studies focus on image restoration to mitigate blur \cite{zhang2021deep, abuolaim2020defocus, ha2024target}. In contrast, other methods directly predict corner locations \cite{zuo2025checkerboard, wu2021highly, du2020correlation, dantas2022automatic, kang2021sparse, zhu2023sub}. In these methods, deep learning models learn to output the pixel coordinates of checkerboard corners via coordinate regression or heatmap estimation, without relying on explicit edge detection or geometric intersection steps. The YOLO family \cite{redmon2016you, chen2023mngnas} has gained particular attention due to its real-time performance and regression-based localisation capability. Variants such as YOLOX and YOLOv8 have demonstrated improved robustness under degraded imaging conditions \cite{wang2022robust, xicai2021binocular, zhou2025checkerboard}. 
\par \setlength\parindent{1em} However, most existing studies evaluate such methods primarily in terms of detection metrics or camera characterisation reprojection error. The effect of feature-localisation robustness on the complete FPP characterisation chain - including projector characterisation, camera-projector stereo consistency, 3D reconstruction, and ultimately dimensional measurement accuracy- has received considerably less attention.
\par \setlength\parindent{1em} Projector characterisation follows a formulation analogous to camera characterisation, treating the projector as an inverse camera \cite{wu2026review}, in which phase measurements establish a correspondence between camera pixels and projector pixel coordinates \cite{zhang2017sub}. In this process, each point on the characterisation board is first mapped from image space to projector space via the phase information, and then associated with its known world coordinates to estimate the projector parameters \cite{Li:14, 6375029}. While these mapping strategies can improve the accuracy of projector coordinate estimation, they remain sensitive to surface reflectivity and low signal-to-noise ratio \cite{rao2018local}. In particular, the sharp changes in reflectivity at the boundaries between black and white checkerboard squares can degrade the captured fringe signals and introduce phase errors \cite{liu2023improved, hu2023phase}. Variations in surface reflectivity may also reduce fringe contrast or cause local underexposure and saturation, further affecting phase estimation \cite{li2023fringe}. Since the absolute phase is used to establish the corresponding projector coordinates, these phase errors can reduce the accuracy of the world-to-projector correspondences and consequently affect projector characterisation.
\par \setlength\parindent{1em} Errors arising from both camera and projector characterisation can propagate into the camera–projector stereo characterisation, affecting the estimation of their relative geometric relationship and consequently introducing systematic errors into 3D reconstruction \cite{zhang2002flexible, Feng2021CalibrationReview}. Since stereo characterisation relies on consistent geometric parameters for both devices, biases in either characterisation can lead to errors in their estimated relative pose. These errors subsequently affect triangulation, degrading reconstruction accuracy in both local geometry and global dimensional measurements \cite{rao2018local}.
\par \setlength\parindent{1em} In this paper, we develop a robust end-to-end characterisation framework for FPP systems operating under non-ideal imaging conditions. The framework combines data-driven checkerboard feature localisation with a centre-based projector correspondence strategy while retaining the conventional geometric camera-projector characterisation model. A YOLOv11 pose-estimation model is used to localise checkerboard corners under clean and degraded imaging conditions. The detected corners are subsequently reordered and supplied to the standard camera characterisation procedure. For projector characterisation, the corner coordinates are used to determine the geometric centres of the white checkerboard squares, and the corresponding absolute phase values are sampled at these locations. This avoids direct phase sampling at black-white reflectivity discontinuities and provides camera-projector correspondences from more uniform reflective regions.

\begin{figure*}[htb]
    \begin{adjustbox}{minipage=\linewidth-4pt,rndframe={width=1pt,color=aa}{5pt}}
        \centering
        \includegraphics[width=0.98\linewidth]{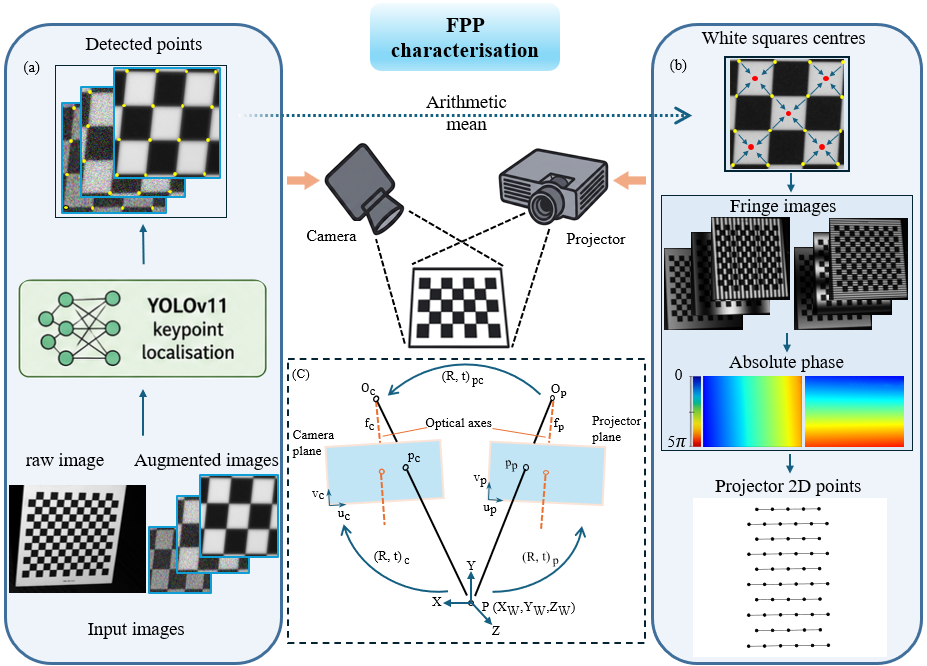}
        \caption{Overview of the proposed FPP characterisation framework. (a) Camera  characterisation: checkerboard corners are localised using the YOLOv11 pose-    estimation model, reordered according to the checkerboard topology, and supplied to the conventional geometric camera-characterisation procedure. (b) Projector characterisation: white-square centres are calculated from the detected checkerboard corners, and absolute phase values sampled at these locations are converted into projector pixel coordinates. (c) Geometric camera-projector model used for camera, projector, and stereo characterisation. The subscripts (c) and (p) denote camera and projector quantities, respectively.}
        \label{fig:methodology2}
    \end{adjustbox}
\end{figure*}

\par \setlength\parindent{1em} The primary focus of the proposed framework is characterisation under degraded imaging conditions, where conventional checkerboard localisation becomes increasingly unreliable; noise-free images are additionally evaluated to provide a reference baseline under favourable conditions. The contributions of this work are threefold. First, we investigate robust learned checkerboard localisation specifically in the context of FPP characterisation and assess its behaviour under both uniform and spatially varying image degradation. Second, we introduce a centre-based projector correspondence strategy in which phase information is sampled at white-square centres derived from the detected checkerboard geometry. Third, rather than evaluating feature detection in isolation, we examine the complete propagation of characterisation performance through camera, projector, and stereo estimation to 3D reconstruction and dimensional measurement. The resulting reconstructions are evaluated against Coordinate Measuring Machine (CMM) reference measurements of physical artefacts, providing both local geometric and global dimensional measures of reconstruction fidelity.

\section*{Materials and methods}

\paragraph{Overview of the proposed characterisation framework} \label{sub: overview}
\hspace*{0.5em} Figure~\ref{fig:methodology2} illustrates the proposed FPP characterisation framework, termed RACE-FPP (Robust AI-assisted Characterisation Enhancement for Fringe Projection Profilometry). The workflow comprises three main stages: camera characterisation, projector characterisation, and camera-projector stereo characterisation, followed by dense 3D reconstruction.
\par \setlength\parindent{1em} For camera characterisation, checkerboard images acquired at multiple poses are first processed using a YOLOv11 pose-estimation model to localise the checkerboard corners. Unlike conventional approaches, such as OpenCV, where corner extraction relies on analytical image processing techniques, RACE-FPP learns corner representations from a diverse training dataset, improving robustness under challenging imaging conditions such as lens blur, noise, and extreme checkerboard orientations.
\par \setlength\parindent{1em} The predicted corner coordinates are subsequently reordered into a consistent checkerboard grid structure before being supplied to the standard camera characterisation procedure based on Zhang’s method. The underlying mathematical characterisation model remains unchanged; the proposed approach replaces only the image feature-localisation stage. Following camera characterisation, the detected corner coordinates are used to compute the centres of the checkerboard white squares. Absolute phase values are then sampled at these locations and converted into projector pixel coordinates, establishing correspondences between world and projector coordinates for projector characterisation. Sampling within the white-square centres avoids the reflectivity discontinuities present at checkerboard corners and is intended to provide more reliable phase information for projector characterisation.
\par \setlength\parindent{1em} The estimated camera and projector parameters are subsequently used to determine the relative rotation and translation between the two devices through stereo characterisation. Finally, the resulting camera-projector geometry, together with the recovered absolute phase maps, is used for dense 3D reconstruction. The performance of the complete framework is assessed at several levels: checkerboard localisation accuracy and detection robustness, camera and projector reprojection errors, camera-projector stereo consistency, and CMM-referenced dimensional accuracy of reconstructed artefacts.

\paragraph{Camera characterisation} \label{sub: cam calib}
\hspace*{0.5em} Camera characterisation establishes the relationship between a 3D point in the target coordinate system, \(\mathbf{P}_w=[X_W,\,Y_W,\,Z_W,\,1]^{\mathsf T}\), and its corresponding 2D location in the camera image, \(\mathbf{p}_c=[u_c,\,v_c,\,1]^{\mathsf T}\), in the homogeneous form. This mapping underpins the subsequent camera-projector characterisation and 3D reconstruction pipeline \cite{zhang2002flexible}. As shown in Figure~\ref{fig:methodology2}(c), the mapping from the world coordinates to the image coordinates can be described using the pinhole model as follows \cite{zhang2017sub, kong2026focus}:

\begin{equation}
\label{eq:mapping}
    \textbf{s}\begin{bmatrix}
         u_c  \\
         v_c  \\
         1
    \end{bmatrix} = \textbf{$K_c$} \begin{bmatrix}
        \textbf{$R_c$}|\textbf{$t_c$}
    \end{bmatrix}
    \begin{bmatrix}
        \textbf{$X_W$}\\
        \textbf{$Y_W$}\\
        \textbf{$Z_W$}\\
        \textbf{1}
    \end{bmatrix},    
\end{equation}
where $s$ is an arbitrary scale factor, \textbf{$K_c$} is the camera intrinsic matrix (including the focal length $f_c$ to the original point $O_c$), and [\textbf{$R_c|t_c$}] is the rotation and translation between the target and the camera coordinate systems. Camera characterisation therefore consists of estimating \textbf{$K_c$}, [\textbf{$R_c|t_c$}] and lens distortion by minimising the reprojection error between the observed 2D image points and their corresponding projected points.
\par \setlength\parindent{1em} Before camera characterisation, the predicted keypoints are reordered into a fixed grid sequence corresponding to the checkerboard topology. Since the point-wise detections are returned according to detection confidence rather than their spatial arrangement, a spatial sorting procedure is applied in which the points are grouped into rows according to their vertical coordinates and subsequently sorted horizontally within each row. Owing to variations in checkerboard orientation between poses, the spatial grouping thresholds in the horizontal and vertical directions are adjusted for individual images to obtain a consistent row-major ordering. The resulting ordered image points are then supplied to the standard planar characterisation procedure based on Zhang's method \cite{zhang2002flexible}. The geometric camera model and parameter-estimation procedure therefore remain unchanged. For comparison, the same characterisation procedure is also performed using corner coordinates obtained with OpenCV's \texttt{findChessboardCornersSB()} implementation. This sector-based detector provides direct sub-pixel corner localisation with improved robustness to image noise and runs faster on large images compared to the conventional \texttt{findChessboardCorners()} implementation \cite{duda2018accurate, hillen2023enhanced}. Consequently, differences between the two camera-characterisation pipelines arise from the image coordinates provided to the common geometric estimation procedure. To evaluate performance under different imaging conditions, three characterisation datasets are considered:
\begin{enumerate}
    \item noise-free images: checkerboard images acquired under favourable imaging conditions, with uniform illumination and minimal visible degradation,
    \item mixed images: a combination of noise-free and degraded checkerboard images. The degraded images include both experimentally observed and synthetically introduced effects and may contain spatially non-uniform degradation (including partial, directional, or full-image defocus) and combinations of blur and image noise, 
    \item noisy images: checkerboard images containing substantial image degradation, including synthetic or experimentally observed noise and blur, used to assess characterisation robustness under challenging acquisition conditions.
\end{enumerate}
\par \setlength\parindent{1em} The reason for these three different sets of images is to demonstrate the broad applicability of RACE-FPP, which yields good characterisation results under various conditions. Each set is processed independently using both RACE-FPP and OpenCV’s checkerboard detection implementation. The evaluation is conducted using two complementary metrics. First, we directly compare the predicted keypoints with the manually annotated reference keypoints described in the \textbf{Annotation strategy} subsection to quantify pixel-level detection accuracy. Second, the reprojection error is used to assess the camera characterisation accuracy, measuring the discrepancy between observed image points and their projections based on estimated parameters. 

\paragraph{Centre-based projector characterisation} \label{sub: proj meth}

\begin{figure*}[!t]
    \begin{adjustbox}{minipage=\linewidth-4pt,rndframe={width=1pt,color=aa}{5pt}}
        \centering
        \includegraphics[width=1\linewidth]{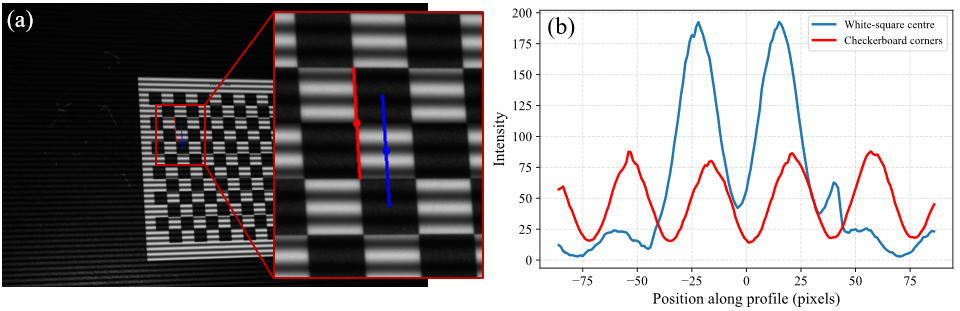}
        \caption{The modulation of the high-frequency horizontal fringes in a sample image: (a) a selected profile region in the image and (b) the intensity modulation in the checkerboard square centre and corners.}
        \label{fig:fringe img}
    \end{adjustbox}
\end{figure*}

\hspace*{0.5em} Similar to the camera, the projector characterisation can be performed based on the pinhole model (Figure~\ref{fig:methodology2}(c)), where the corresponding 2D projector points are represented as $p_p$. Assuming the projector behaves as an inverse imaging system that follows the same pinhole model, it can be characterised using a planar checkerboard pattern in accordance with Zhang and Huang’s method \cite{zhang2006novel}. Unlike a camera, however, the projector does not directly observe the calibration target. Correspondences between points observed by the camera and coordinates in the projector image plane must therefore be established indirectly from the projected fringe patterns, as shown in Equation~\ref{eq: projector points}.

\par \setlength\parindent{1em} Two orthogonal multi-frequency sinusoidal fringe patterns are sequentially projected onto the checkerboard. The fringes are projected horizontally ($\mathrm{H}$) and vertically ($\mathrm{V}$) to obtain the projector pixel coordinates $(u_p, v_p)$ corresponding to each camera pixel. The use of multiple fringe frequencies enables reliable phase unwrapping and recovery of the absolute phase map. Let $(u_c, v_c)$ denote a point in the camera image coordinates and $(u_p, v_p)$ the corresponding projector coordinates. For each fringes orientation $d \in \{\mathrm{V}, \mathrm{H}\}$, an $N$-step phase-shifting sequence, $\delta_i$, with known phase shifts is projected and captured:

\begin{equation}
\label{eq: phase shifting}
\delta_i = \frac{2\pi(i-1)}{N}, \quad i = 1, \ldots, N.
\end{equation}

The recorded intensity at each pixel is given by:

\begin{equation}
\begin{aligned}
I^{(d)}_i(u_c,v_c)
&= A^{(d)}(u_c,v_c) + B^{(d)}(u_c,v_c) \\
&\quad \times \cos\big(\phi^{(d)}(u_c,v_c)+\delta_i\big).
\end{aligned}
\label{eq:intensity}
\end{equation}

where $A^{(d)}(u_c, v_c)$ and $B^{(d)}(u_c, v_c)$ represent the background intensity and modulation amplitude, respectively, and $\phi^{(d)}(u_c, v_c)$ is the phase to be recovered. Vertical fringes ($d = \mathrm{V}$) encode the projector $u_p$ coordinate, while horizontal fringes ($d = \mathrm{H}$) encode the projector $v_p$ coordinate. The wrapped phase (Equation~\ref{eq: wrapped phase}) is first obtained from Equation~\ref{eq:intensity} using the phase-shifting demodulation via the least-squares method and, subsequently, unwrapped to yield the absolute phase ($\Phi^{(d)}(u_c, v_c)$) by adding an integer that represents the "fringe order" (the number of cycles that the phase has wrapped), as shown in Equation~\ref{eq: unwrapped phase}:

\begin{equation}
    \label{eq: wrapped phase}
    \phi^{(d)}(u_c, v_c) = tan^{-1}\frac{\sum_{i=1}^{N} I^{(d)}_i(u_c, v_c) sin(\delta_i)}{\sum_{i=1}^{N} I^{(d)}_i(u_c, v_c) cos(\delta_i)}
\end{equation}

\begin{equation}
    \label{eq: unwrapped phase}
    \Phi^{(d)}(u_c, v_c) = \phi^{(d)}(u_c, v_c) + 2\pi K(u_c, v_c)
\end{equation}

The projector coordinates are then computed as:
\begin{equation}
\label{eq: projector points}
u_p = \frac{\Phi^{(\mathrm{V})}(u_c, v_c)}{2\pi f_{\mathrm{V}}}, \quad
v_p = \frac{\Phi^{(\mathrm{H})}(u_c, v_c)}{2\pi f_{\mathrm{H}}},
\end{equation}

where $f_{\mathrm{V}}$ and $f_{\mathrm{H}}$ denote the spatial carrier frequencies of the vertical and horizontal fringe patterns, respectively.

\par \setlength\parindent{1em} In contrast to traditional methods that assign phase values to the checkerboard’s corner points, this work associates the phase information with the centres of the white squares. This strategy enhances the robustness of the phase data because the white areas exhibit smoother and more uniform fringe modulation, while the corner regions include grey-level transitions that introduce phase ambiguities, as can be seen in Figure~\ref{fig:fringe img}. In conventional checkerboard-based projector characterisation, the phase values used to determine projector coordinates are commonly evaluated at the checkerboard corner locations. These positions coincide with abrupt transitions between black and white regions of the target. Such reflectivity discontinuities can reduce local fringe modulation and introduce underexposure, saturation, or low signal-to-noise ratio, thereby increasing uncertainty in the recovered phase \cite{liu2023improved, rao2018local}.

\par \setlength\parindent{1em} To reduce sensitivity to these effects, the proposed framework associates the projector phase information with the centres of the white checkerboard squares rather than directly with the checkerboard corners. As illustrated in  Figure~\ref{fig:fringe img}, the white-square interiors provide more spatially uniform reflective regions and exhibit stronger fringe modulation than the black-white transition regions.

\par \setlength\parindent{1em} The square centre coordinates are derived from the checkerboard corner points detected in the previous stage. The centre position of the white square is computed as the arithmetic mean of the coordinates of its four surrounding corners. Since the detected corner locations are usually sub-pixel values, the resulting centre does not usually coincide with a discrete pixel location in the phase map. Consequently, the phase value at the square centre is not directly available and must be estimated from the surrounding phase-map pixels. To obtain a sub-pixel phase estimate, bilinear interpolation is applied. For a centre located at $(x_s, y_s)$, the interpolated phase $\Phi(x_s, y_s)$ is computed as:

\begin{equation}
\label{eq: centre phase}
\Phi(x_s, y_s) = \sum_{i=1}^{2} \sum_{j=1}^{2} w_{ij} \, \Phi(x_i, y_j),
\end{equation}

where $(x_i, y_j)$ denote the four neighbouring phase-map pixels surrounding the centre location, and $w_{ij}$ are interpolation weights determined by the relative distances between the centre location and these pixels, satisfying $\sum_{i,j} w_{ij} = 1$. Neighbouring pixels closer to the centre are assigned larger interpolation weights than those farther away.
\par \setlength\parindent{1em} The projector characterisation followed the same image grouping strategy adopted for camera characterisation, using the clear, mixed, and noisy datasets. For each set, intrinsic and extrinsic parameters are estimated independently. The reprojection error of each configuration is used to compare the proposed square-centre approach against the conventional corner-based method implemented in OpenCV.
\par \setlength\parindent{1em} Following the individual characterisation of the camera and projector, stereo characterisation is performed to estimate the relative transformation between them. This step determines the rotation matrix and translation vector (R, t)$_{pc}$ that define the projector pose relative to the camera, which, in this work, serves as the reference coordinate system for subsequent reconstruction. These parameters are essential for triangulating 3D points during reconstruction.
\par \setlength\parindent{1em} Using the estimated camera, projector, and stereo characterisation parameters together with the recovered absolute phase maps, 3D reconstruction is then performed. The characterisation parameters define the geometric relationship between the devices, while the absolute phase establishes correspondences between camera pixels and projector coordinates. The resulting reconstructed geometries are compared against reference measurements obtained using a Coordinate Measuring Machine (CMM; Mitutoyo Crysta Apex S7106)  and analysed in terms of dimensional and geometric deviations, following established approaches for the metrological validation of optical 3D measurement systems \cite{catalucci2020measurement, shaheen2021characterisation}. This provides a practical validation of the complete characterisation pipeline.

\section*{Deep learning model}
\noindent
In this work, the checkerboard corner detection problem is reformulated as a keypoint detection task, where each corner is treated as an independent target. A YOLOv11 pose estimation model is adopted as the base architecture due to its efficient inference and improved localisation capability compared to earlier YOLO variants \cite{khanam2024yolov11}. YOLOv11 follows a unified backbone–neck–head architecture. The backbone extracts visual features from the input images at multiple scales via its internal layers; the neck combines information from different resolutions; and the pose head predicts the locations of the checkerboard corner keypoints. This design enables robust localisation under varying image conditions and checkerboard poses.

\paragraph{Dataset preparation}
\hspace*{0.5em} The keypoint detection of the deep learning model heavily relies on the availability of an accurate labelled training dataset. In this work, the dataset includes both original images and augmented variations to increase robustness to different imaging conditions such as blur and noise.
\par
\begin{figure*}[htb]
    \begin{adjustbox}{minipage=\linewidth-4pt,rndframe={width=1pt,color=aa}{5pt}}
        \centering
        \includegraphics[scale=0.65]{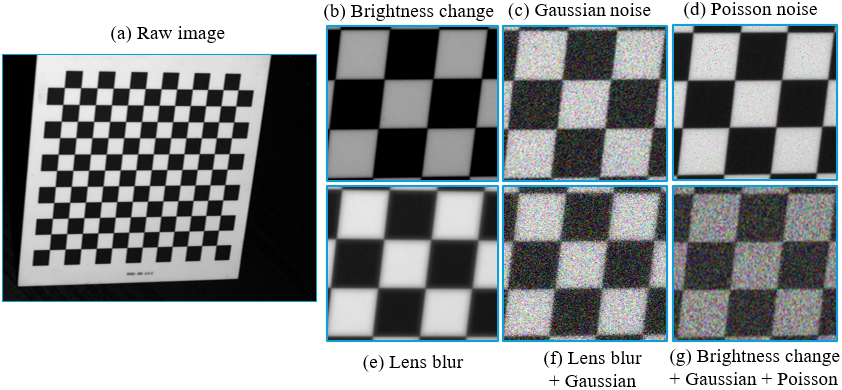}
        \caption{Dataset samples used for training with: (a) the original raw image and (b) - (g) the added noise types for the same image in a unified area.}
        \label{fig:noise types}
    \end{adjustbox}
\end{figure*}

 \setlength\parindent{1em} High-resolution images (3000 $\times$ 2000 pixels) of a characterisation target are captured using a fringe projection system to train the deep learning model. The characterisation plate is a 12 $\times$ 13 checkerboard pattern, resulting in 132 interior corner points with a square size of 6 mm. The checkerboard is captured 200 times in different positions and orientations relative to the camera to ensure geometric diversity. The images were taken using an FPP system, developed by Taraz Metrology Ltd, with a Digital Light Processing (DLP) projector that emits sinusoidal fringe patterns generated by a bespoke high-performance computer (Windows 10, 64-bit architecture, and 32 GB RAM). The imaging camera employs a 16 mm focal length lens with an F/1.8 aperture and provides a field of view (FOV) of approximately 255 mm $\times$ 174 mm. To improve training generalisation and better reflect realistic FPP imaging conditions, the dataset was augmented by synthetically applying a range of photometric and blur-related degradations to the original checkerboard images.  These included changes in brightness and contrast, gamma adjustment, Poisson and Gaussian noise, and both full-frame and directional defocus blur. Several augmentations were also applied in combination, as shown in Figure~\ref{fig:noise types}.

\paragraph{Annotation strategy}
\hspace*{0.5em} Each checkerboard image was annotated using LabelMe \cite{wada2016labelme}, an open-source image annotation tool that enables annotations to be created and stored locally. Manual annotation was adopted because no automatic corner detection method could be assumed to provide error-free ground-truth labels under the challenging imaging conditions considered in this work. The annotations were performed at high image magnification and subsequently reviewed to maintain consistent corner placement and ordering. In the adopted point-wise formulation, each internal checkerboard corner is treated as an independent instance; consequently, each label file contains 132 annotation lines, one for each of the 132 internal corners. Each annotation consists of a bounding box and its associated keypoint coordinate in the form: class  $x_{box}$ $y_{box}$  $w_{box}$ $h_{box}$  $x_k$ $y_k$  $v_k$,  where class is a number representing the object class (in our case, it is set to 0 as we only have one class, the checkerboard), $x_{box}, y_{box}, w_{box}, h_{box}$ denote the normalised centre and size of the instance's bounding box, and each triplet ($x_k, y_k, v_k$) represents the normalised keypoint coordinates and their visibility flag, where $v_k \in \{0,1,2\}$. A value of $v_k$=2 indicates the keypoint is visible, $v_k$=1 means labelled but occluded, and $v_k$=0 marks absent points. Most of our checkerboard corners remain fully visible within the FOV and are therefore assigned a visibility value of 2, following the standard YOLO pose annotation convention \cite{saha2025deep}. For a few captured poses, however, portions of the checkerboard extend outside the image boundaries, resulting in invisible points. Maintaining a consistent annotation structure across all 132 corners is essential to preserving stable point correspondence during training.
\par \setlength\parindent{1em} Unlike conventional checkerboard detection, which requires successful identification of the complete predefined corner pattern for an image to be accepted, the proposed approach adapts the target representation by treating each checkerboard corner as an independent detection target, without modifying the underlying network architecture. Consequently, local degradation affecting individual corners does not necessarily invalidate detections obtained from the remaining regions of the checkerboard, while confidence scores provide an additional measure for filtering uncertain predictions. The practical effect of this point-wise formulation on the number of successfully detected characterisation images is evaluated later in Table~\ref{tab:detection_cam_rpe}.

\paragraph{Model training and testing}
\hspace*{0.5em} The YOLOv11 pose estimation network is trained using the annotated checkerboard dataset described in the previous section. The dataset is divided into 80\% for training and 20\% for validation, ensuring that both subsets contain data spanning the full range of imaging conditions. Instead of dedicating a percentage of the dataset to testing, the trained model is later evaluated on real characterisation images against ground-truth point correspondences. The training is performed using the YOLOv11 framework built on PyTorch with an NVIDIA Quadro P5000 Graphics Processing Unit (GPU).
\par \setlength\parindent{1em} A higher input resolution (1440 $\times$ 960 pixels) than the commonly used YOLO settings (usually 640 $\times$ 640 pixels \cite{sapkota2026ultralyticsyoloevolutionoverview}) is selected to preserve better checkerboard corner details required for accurate characterisation, while remaining within the available GPU memory constraints. The training hyperparameters are set to 200 epochs with a batch size of four at a 0.001 learning rate, decaying linearly over the training schedule. Hyperparameters are empirically tuned based on validation performance. AdamW is used as the optimiser. Training employed early stopping with a patience of 20 epochs, terminating when the validation performance no longer improved, thereby preventing overfitting.

\begin{figure*}[!t]
    \begin{adjustbox}{minipage=\linewidth-4pt,rndframe={width=1pt,color=aa}{5pt}}
        \centering
        \includegraphics[width=0.9\linewidth]{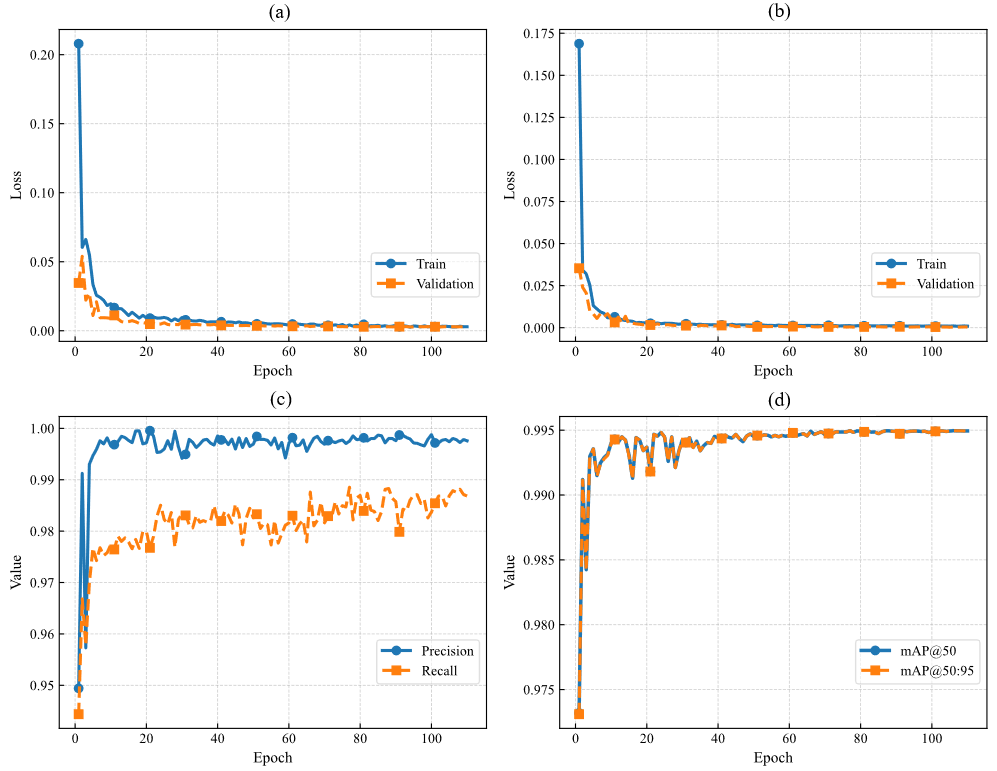}
        \caption{YOLOv11 pose estimation model metrics: (a) and (b) are the pose and keypoint objectness losses for training optimisation, respectively, and (c) and (d) are precision and recall,  and Mean Average Precision graphs, respectively, for performance evaluation.}
        \label{fig:model metrics}
    \end{adjustbox}
\end{figure*}

\par \setlength\parindent{1em} The model training is optimised using five loss components associated with bounding box localisation, keypoint localisation, keypoint objectness, classification, and distribution focal loss. Since the present application focuses on checkerboard corner localisation, particular attention is given to the pose and keypoint objectness losses. During training, model performance is monitored using the validation losses together with precision, recall, and mean average precision (mAP) \cite{sapkota2026ultralyticsyoloevolutionoverview}. Precision measures the proportion of predicted detections that are correct, whereas recall measures the proportion of ground-truth instances that are successfully detected. For keypoint evaluation, mAP summarises the precision–recall performance based on intersection-over-union (IoU), with mAP@0.5 evaluated at an IoU threshold of 0.5 and mAP@0.5:0.95 averaged over IoU thresholds from 0.5 to 0.95. Once trained, the model predictions are evaluated against manually annotated reference keypoints prior to characterisation. The reference keypoints are obtained using the annotation procedure described in the \textbf{Annotation strategy} subsection. The predictions are also compared with those obtained using OpenCV's conventional checkerboard detection algorithm.

\section*{Results and discussion}
\noindent
\paragraph{Keypoint detection performance}
\hspace*{0.5em} The training behaviour of the corner-detection model is first examined to assess convergence before evaluating its localisation performance on the independent characterisation datasets.  Figure~\ref{fig:model metrics} represents the evolution of the training and validation pose and keyobjectness losses (Figure~\ref{fig:model metrics}(a) and (b)),  precision and recall (Figure~\ref{fig:model metrics}(c)), and mean average precision (mAP) (Figure~\ref{fig:model metrics}(d)). Training terminated after 110 epochs (25.231 hours) through the early-stopping criterion, when no further improvement was observed over the specified patience period. \begin{table*}[!t]
\centering
%\scriptsize
\caption{Percentage of correct keypoints (PCK) at different thresholds for RACE-FPP and OpenCV detections.}
\label{tab:pck_comparison}

%\resizebox{\linewidth}{!}{%
\begin{tabular}{lcccccccc}
\toprule
\multirow{2}{*}{Image dataset} & 
\multicolumn{4}{c}{RACE-FPP} & 
\multicolumn{4}{c}{OpenCV} \\
\cmidrule(lr){2-5} \cmidrule(lr){6-9}
& PCK@0.5 & PCK@1 & PCK@5 & PCK@10 
& PCK@0.5 & PCK@1 & PCK@5 & PCK@10 \\
\midrule

Noise-free  & 19.70 & 100.00 & 100.00 & 100.00 & 12.12 & 100.00 & 100.00 & 100.00\\
Mixed images & 16.67 & 58.33 & 100.00 & 100.00 & 16.67 & 49.24 & 96.21 & 97.73\\
Noisy images & 3.03 & 19.70 & 99.24 & 100.00 & 5.30 & 15.15 & 77.27 & 87.88\\

\bottomrule
\end{tabular}
%}
\end{table*}
\par \setlength\parindent{1em} As seen in Figure~\ref{fig:model metrics}(a) and (b), the training losses for pose and keypoint objectness decreased smoothly from approximately 0.208 and 0.17 to 2.93 $\times 10^{-3}$ and 9.4 $\times 10^{-4}$, respectively, before converging toward stable values as the model progressively learned more accurate corner localisation patterns from the training data. The corresponding validation curves exhibit similar convergence behaviour and remain close to the training curves, with no evident divergence as training progresses. These losses' trends indicate stable optimisation of the model on both the training and validation datasets.
\par \setlength\parindent{1em} The detection metrics in Figure~\ref{fig:model metrics}(c) show a rapid improvement during the initial training epochs before stabilising, with final precision and recall values of approximately 99.8\% and 98.7\%, respectively. The high precision indicates that few predicted corner instances are false detections, while the recall indicates that the majority of reference corner instances are successfully detected. Figure~\ref{fig:model metrics}(d) further shows that both mAP@0.5 and mAP@0.5:0.95 converge to approximately 99.5\%. The close agreement between the two mAP measures indicates that the high detection performance is maintained across the evaluated IoU thresholds.
\par \setlength\parindent{1em} To complement these global metrics, the network’s spatial accuracy was further examined using the Percentage of Correct Keypoints (PCK), a standard metric in keypoint detection and pose estimation tasks \cite{yu2025yolo}. PCK@$t$ denotes the percentage of detected keypoints whose localisation error is within a threshold $t$ (in pixels):
\begin{equation}
\label{eq: PCK}
    \mathrm{PCK@}t = \frac{N_{\text{correct}}}{N_{\text{total}}} \times 100\%,
\end{equation}
where $N_{\text{correct}}$ is the number of keypoints with localisation error below the threshold $t$. In this study, thresholds of five pixels and 10 pixels (px) were initially adopted to represent moderate and relaxed localisation tolerances relative to the image resolution, enabling the detector's robustness to be evaluated under increasingly degraded imaging conditions.
\par \setlength\parindent{1em} The resulting PCK@5 px and PCK@10 px values reached 100\% across the noise-free and mixed datasets and over 99\% for noisy images, as illustrated in Table~\ref{tab:pck_comparison}, indicating that almost all detected corners were localised within five pixels of the reference annotations. This outcome aligns with the high precision and recall trends observed during training, reinforcing the model’s spatial consistency and prediction reliability. At stricter localisation thresholds, RACE-FPP achieves comparable or higher PCK values than OpenCV at 1 px, while both methods exhibit relatively low PCK at 0.5 px. This indicates that, when corners are successfully detected, the localisation performance of the two methods becomes more comparable at demanding sub-pixel thresholds. However, RACE-FPP maintains successful detection of the complete checkerboard across the investigated image conditions, as will be seen in the \textbf{Camera characterisation} subsection, providing a more consistent set of image observations for the subsequent characterisation stages.

\begin{figure*}[h!]
    \begin{adjustbox}{minipage=\linewidth-4pt,rndframe={width=1pt,color=aa}{5pt}}
        \centering
        \includegraphics[width=0.9\linewidth]{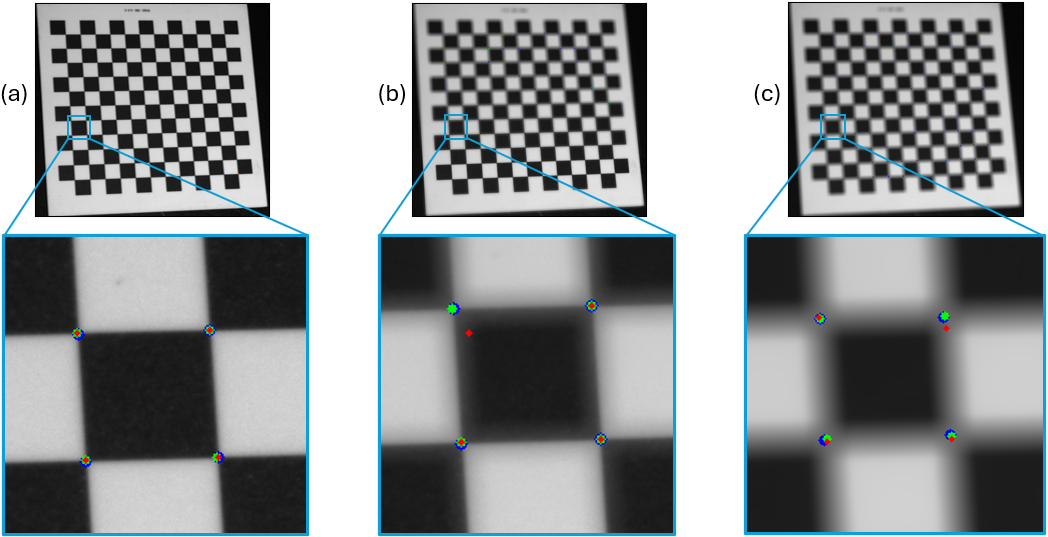}
        \caption{Comparison of checkerboard corner localisation using the proposed method (green), OpenCV (red), and manually annotated ground truth (blue) under three representative image qualities: (a) noise-free image, (b) partial horizontal defocus, and (c) severe full-image defocus.}
        \label{fig:superimposition}
    \end{adjustbox}
\end{figure*}

\paragraph{Corner localisation accuracy and robustness} \label{sub: comparative analysis}
\hspace*{0.5em} To further evaluate the spatial accuracy of the proposed deep learning detector, its keypoint predictions are compared against the outputs of OpenCV’s \texttt{findChessboardCornersSB()} function and the manually annotated ground truth. The analysis uses three representative test images acquired with a fixed checkerboard pose, representing different image qualities -noise-free, partial defocus, and full defocus images- to assess detection robustness under different characterisation scenarios. Figure~\ref{fig:superimposition} presents the resulting corner locations superimposed on the corresponding checkerboard images. RACE-FPP predictions are marked in green, the OpenCV detections in red, and the ground truth (GT) corners in blue. The zoomed-in region is kept consistent across all three images to facilitate a direct comparison of corner localisation under different image degradations. The partially defocused image is particularly representative of practical characterisation conditions, where variations in depth and viewing geometry result in focused and defocused regions within the same image. Visual inspection reveals that the proposed model reliably identifies all 132 corner points with accurate spatial alignment for all the images, while the OpenCV method exhibits localisation deviations, particularly in degraded regions. Inaccurate detections are observed for OpenCV under non-ideal conditions in Figure~\ref{fig:superimposition} (b) and (c), as will be discussed further in the following subsection. Interestingly, the OpenCV detections in the partially defocused image exhibit localisation errors comparable to, and in some locations greater than, those observed under full-image defocus. This behaviour may be attributed to the transition region containing spatially varying blur, producing locally inconsistent image gradients on which conventional gradient-based corner detectors rely. In contrast, the fully defocused image exhibits more uniform blur characteristics, despite the overall loss of sharpness. Similar challenges associated with spatially varying defocus have been reported in the literature on blur estimation and image feature analysis \cite{zhuo2011defocus, shi2015just, deng2025multi}.

\par
\begin{figure*}[h!]
    \begin{adjustbox}{minipage=\linewidth-4pt,rndframe={width=1pt,color=aa}{5pt}}
        \centering
        \includegraphics[width=1\linewidth]{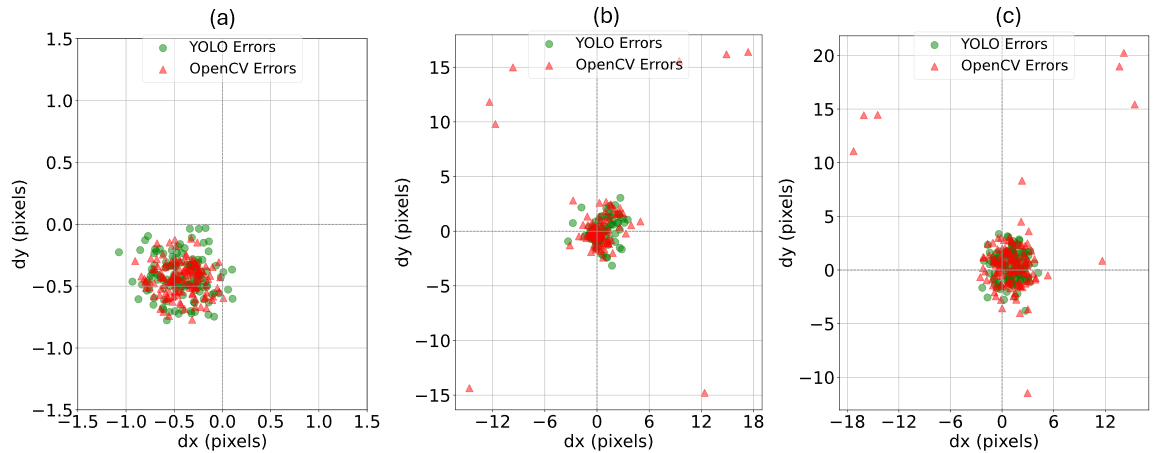}
        \caption{Spatial distribution of the directional localisation errors (\(D_x, D_y\)) relative to the manually annotated ground truth for the same representative checkerboard images shown in Figure~\ref{fig:superimposition}.}
        \label{fig:Euclidean}
    \end{adjustbox}
\end{figure*}

 \setlength\parindent{1em}  For a quantitative assessment, the per-point localisation (Euclidean) error ($e_i$), defined in Equation~\ref{eq: euclidean}, is computed for each detected corner, providing a direct measure of geometric accuracy in image-pixel coordinates \cite{yu2025yolo}. For each checkerboard corner, the predicted keypoint is matched to its corresponding ground-truth point using the predefined grid ordering, and the resulting localisation error directional components are visualised in Figure~\ref{fig:Euclidean} for the representative checkerboard images presented in Figure~\ref{fig:superimposition}. The localisation errors are then summarised by their mean and standard deviation over all 132 checkerboard corners and reported in Table~\ref{tab:error_stats}, along with Euclidean magnitudes. The error statistics for RACE-FPP and the OpenCV detections in the horizontal and vertical directions, together with the overall mean Euclidean error, are presented.
\begin{equation}
\label{eq: euclidean}
   e_i = \sqrt{(x_i^{pred}-x_i^{gt})^2 + (y_i^{pred}-y_i^{gt})^2} , 
\end{equation}

where $x_i^{pred}$, $y_i^{pred}$, $x_i^{gt}$ and $y_i^{gt}$ indicate the $i^{th}$ point's coordinates from the predictions and the ground truth values, respectively.

\par \setlength\parindent{1em} Figure~\ref{fig:Euclidean} illustrates the spatial distribution of localisation errors for both methods across the same images in Figure~\ref{fig:superimposition}. RACE-FPP errors are labeled in green, and the OpenCV ones are in red. Figure~\ref{fig:Euclidean}(a) shows compact localisation-error distributions for both methods under noise-free conditions, with most errors remaining below one pixel. Both distributions exhibit a similar systematic offset from the origin, with mean directional errors of approximately \(-0.4\) px in both axes (Table~\ref{tab:error_stats}). The close agreement between the offsets observed for RACE-FPP and OpenCV indicates that this displacement is common to both methods relative to the manually annotated reference, rather than being specific to the proposed detector. In contrast, the OpenCV detections exhibit larger variations and noticeable outliers reaching almost 18 pixels when blurry conditions exist in Figure~\ref{fig:Euclidean}(b) and (c). The largest OpenCV errors are concentrated in a small number of corner points rather than being uniformly distributed across the checkerboard, indicating that localisation failures occur locally in degraded image regions. These outliers correspond to incorrectly localised corner points, which typically occur in regions affected by image degradation. Compared with OpenCV, the proposed method exhibits a tighter clustering of localisation errors without the presence of extreme outliers in Figure~\ref{fig:Euclidean}(b) and (c), indicating that the proposed detector maintains consistently low localisation errors even when conventional gradient-based corner detection produces isolated large localisation failures.
\par \setlength\parindent{1em} These observations are quantitatively confirmed by the error statistics reported in Table~\ref{tab:error_stats}, where RACE-FPP consistently exhibits lower mean localisation errors and substantially smaller standard deviations than OpenCV, particularly under partial and severe defocus.

\begin{table*}[htbp]
\centering
%\scriptsize  % smallest recommended font
\caption{Mean and standard deviation (std) of the directional (\(D_x, D_y\)) and Euclidean localisation errors in pixels (px) relative to the manually annotated ground truth (GT).}
\label{tab:error_stats}

%\resizebox{\linewidth}{!}{%
\begin{tabular}{lcccccccccccc}
\toprule
\multirow{3}{*}{Error} & 
\multicolumn{6}{c}{RACE-FPP} &
\multicolumn{6}{c}{OpenCV} \\
\cmidrule(lr){2-7} \cmidrule(lr){8-13}
& \multicolumn{2}{c}{Noise-free} 
& \multicolumn{2}{c}{partial defocus} 
& \multicolumn{2}{c}{full defocus} 
& \multicolumn{2}{c}{Noise-free} 
& \multicolumn{2}{c}{partial defocus} 
& \multicolumn{2}{c}{full defocus} \\
\cmidrule(lr){2-3} \cmidrule(lr){4-5} \cmidrule(lr){6-7}
\cmidrule(lr){8-9} \cmidrule(lr){10-11} \cmidrule(lr){12-13}
& Mean & std & Mean & std & Mean & std & Mean & std & Mean & std & Mean & std \\
\midrule

Dx 
& -0.42 & 0.23 & -0.28 & 1.17 & 1.02 & 1.32
& -0.43 & 0.20 & 0.17 & 3.41 & 1.00 & 3.71\\

Dy
& -0.42 & 0.16 &  -0.12 & 0.96 & 0.34 & 1.36
& -0.44 & 0.13 &  0.34 & 3.65 & 1.00 & 3.84\\

Euclidean
& 0.63 & 0.19 & 1.22 & 0.94 & 2.00 & 0.86
& 0.64 & 0.16 & 2.27 & 4.47 & 3.30 & 4.43\\

\bottomrule
\end{tabular}
%}
\end{table*}

\par \setlength\parindent{1em} The trends observed qualitatively in Figures~\ref{fig:superimposition} and~\ref{fig:Euclidean} are quantitatively supported by the error statistics reported in Table~\ref{tab:error_stats}. Under noise-free imaging conditions, both methods exhibit similar localisation performance, with mean Euclidean errors of 0.63 px and 0.64 px for RACE-FPP and OpenCV, respectively. The corresponding mean directional errors are also nearly identical, with \(D_x=-0.42\) px and \(D_y=-0.42\) px for RACE-FPP and \(D_x=-0.43\) px and \(D_y=-0.44\) px for OpenCV, confirming the common offset observed in Figure~\ref{fig:Euclidean}(a). As the image degradation increases from partial to full defocus, the localisation error increases for both methods. However, RACE-FPP consistently maintains lower mean Euclidean errors of 1.22 px and 2.00 px than OpenCV, compared with 2.27 px and 3.30 px for OpenCV. Moreover, the proposed method exhibits substantially smaller standard deviations under both degraded conditions (0.94 px and 0.86 px) than OpenCV (4.47 px and 4.43 px). This indicates that, beyond achieving a lower average localisation error, the proposed detector provides considerably more consistent corner localisation with far fewer large localisation failures. These observations are consistent with the tighter error distributions and reduced number of outliers shown in Figure~\ref{fig:Euclidean}, confirming that RACE-FPP maintains stable localisation performance even under challenging imaging conditions.

\begin{table*}[!t]
\centering
%\scriptsize
\caption{Detection rates and reprojection error (RPE) in pixels (px) for camera characterisation using RACE-FPP and OpenCV pipelines.}
\label{tab:detection_cam_rpe}

\resizebox{\linewidth}{!}{%
\begin{tabular}{lcccc}
\toprule
\textbf{Dataset} & 
\textbf{Images detected (RACE-FPP)} & 
\textbf{RPE (RACE-FPP)} & 
\textbf{Images detected (OpenCV)} & 
\textbf{RPE (OpenCV)} \\
\midrule

Noise-free       & 16/16 & 0.2281 & 16/16 & 0.0806 \\
Mixed images     & 16/16 & 0.2591 & 10/16 & 1.2365 \\
Noisy images     & 16/16 & 0.4500 & 5/16  & 2.8514 \\

\bottomrule
\end{tabular}
}
\end{table*}

\paragraph{Camera characterisation}\label{sub: camera results}
\hspace*{0.5em}To assess the impact of the proposed model on camera characterisation accuracy, three separate characterisations are performed using (i) noise-free images, (ii) a mixed set containing noise-free and noisy images, and (iii) noisy images only. Each characterisation is executed twice: once using RACE-FPP predicted keypoints, and the other using the points detected by OpenCV’s \texttt{findChessboardCornersSB()}. The characterisation follows the standard pinhole camera model with radial and tangential distortion terms \cite{zhang2002flexible}. The root mean square reprojection error (RPE), in Equation~\ref{eq: rpe}, is used to quantify the overall geometric accuracy.

\begin{figure*}[!t]
    \begin{adjustbox}{minipage=\linewidth-4pt,rndframe={width=1pt,color=aa}{5pt}}
        \centering
        \includegraphics[width=0.7\linewidth]{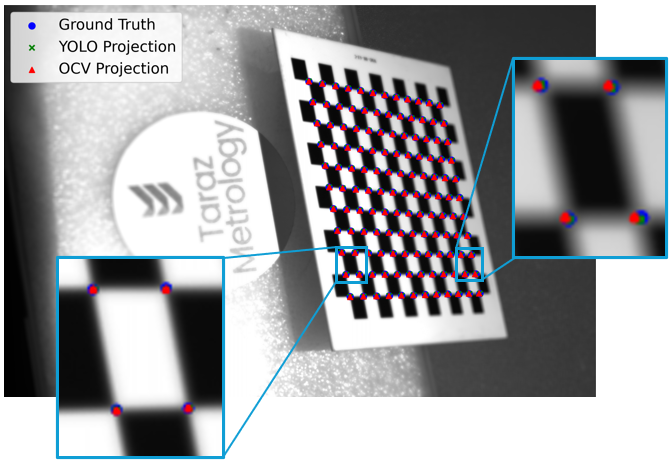}
        \caption{Visualisation of projected 3D points using the estimated camera intrinsic and extrinsic parameters. Areas of different defocus within the image are zoomed in.}
        \label{fig:visualisation}
    \end{adjustbox}
\end{figure*}

\begin{equation}
\label{eq: rpe}
    \mathrm{RPE} = \sqrt{\frac{1}{K}\sum_{j=1}^K\frac{1}{N} \sum_{i=1}^N(p_i^{\text{det}}-p_i^{\mathrm{est}} )^2},
\end{equation}
where $p_i^{\text{det}}$ and $p_i^{\mathrm{est}}$ are the coordinates of the $i^{th}$ point in the $j^{th}$ image of the detected image point (YOLO prediction or OpenCV detection) and the corresponding reprojected point obtained using the estimated characterisation parameters, respectively; $N$ and $K$ are the total number of points and images, respectively. The resulting errors for all three datasets are summarised in Table~\ref{tab:detection_cam_rpe}.

\par \setlength\parindent{1em} In addition to reprojection accuracy, the number of characterisation images successfully processed was evaluated, as a complete set of checkerboard correspondences is required for each pose included in the camera characterisation. RACE-FPP successfully provided the required correspondences for all 16 images in each dataset. The OpenCV baseline also processed all 16 noise-free images, but this number decreased to 10/16 for the mixed dataset and 5/16 for the noisy dataset. Thus, the effect of image degradation was observed not only in the localisation accuracy of the detected corners, but also in the number of poses available for characterisation. Moreover, some of the degraded images successfully accepted by OpenCV exhibited comparatively large per-image reprojection errors. For example, the maximum per-image root mean square (RMS) reprojection error increased from 0.124~px in the noise-free dataset to 3.801~px and 4.788~px in the mixed and noisy datasets, respectively. In contrast, RACE-FPP retained all characterisation poses across the three imaging conditions, demonstrating greater robustness to the imposed image degradation.

\par \setlength\parindent{1em} The reprojection errors for both methods were calculated using the same OpenCV camera characterisation routine and evaluation procedure. Under noise-free conditions, OpenCV achieves a lower global RMS reprojection error of 0.0806~px, compared with 0.2281~px for RACE-FPP. This difference can be attributed, at least in part, to the different localisation strategies. The OpenCV checkerboard detector is specifically designed for geometric corner localisation and provides sub-pixel corner estimates, which is advantageous when the checkerboard edges and corner transitions are clearly resolved. In contrast, RACE-FPP estimates corner locations through a learned keypoint representation trained across a broad range of image degradation conditions. Consequently, under favourable noise-free conditions, the specialised OpenCV detector provides more precise corner localisation, while both methods retain sub-pixel reprojection accuracy. RACE-FPP is primarily developed to address corner localisation under degraded FPP imaging conditions, for which conventional checkerboard detection becomes unreliable. As the imaging conditions deteriorate, however, the RACE-FPP characterisations yield global RMS errors of 0.2591~px and 0.4500~px for the mixed and noisy datasets, respectively, whereas the corresponding OpenCV errors increase to 1.2365~px and 2.8514~px. These values should be interpreted together with the detection success rates: the OpenCV errors under degraded conditions are calculated only from the successfully detected subsets (10/16 and 5/16 poses), whereas RACE-FPP uses all 16 poses in each case. Consequently, the degraded-condition reprojection errors do not constitute a point-for-point comparison over identical pose subsets; rather, they characterise the calibration obtained from the correspondences that each method successfully provides. Taken together, the pose-retention rate and reprojection results show a trade-off between the higher precision of OpenCV under noise-free conditions and the greater robustness of RACE-FPP as image quality deteriorates.

\begin{table}[!t]
\centering
%\scriptsize
\small
\setlength{\tabcolsep}{5pt}
\renewcommand{\arraystretch}{1.75}
\caption{Estimated camera intrinsic parameters obtained using the RACE-FPP and OpenCV pipeline for the three image datasets.}
\label{tab:camera_intrinsics}

\resizebox{\linewidth}{!}{%
\begin{tabular}{llcccc}
\toprule
\textbf{Method} & \textbf{Dataset} &
$f_x \times 10^3$ px & $f_y \times 10^3$ px & $c_x \times 10^3$ px & $c_y \times 10^3$ px \\
\midrule

\multirow{3}{*}{RACE-FPP}
& Noise-free     & 6.708 & 6.705 & 1.432 & 0.910\\
& Mixed images   & 6.733 & 6.728 & 1.429 & 0.896\\
& Noisy images   & 6.714 & 6.711 & 1.420 & 0.903\\
\midrule

\multirow{3}{*}{OpenCV}
& Noise-free     & 6.712 & 6.711 & 1.446 & 0.913\\
& Mixed images   & 6.597 & 6.634 & 1.397 & 1.018\\
& Noisy images   & 6.352 & 6.472 & 1.293 & 1.207\\

\bottomrule
\end{tabular}
}
\end{table}

\par \setlength\parindent{1em} Figure~\ref{fig:visualisation} provides a visual illustration of the projection of the 3D points onto the checkerboard. An image with a tilted checkerboard pose and partial defocus is selected as an example, as it reflects realistic FPP imaging conditions. The figure displays the projected points using the parameter matrices estimated from the two pipelines: RACE-FPP is indicated by green crosses, and OpenCV by red triangles. The reference points, manually annotated using the LabelMe image annotation tool, are shown as blue circles to enable a comprehensive assessment, clearly highlighting the deviation of the OpenCV points in the blurred regions of the image.

\par \setlength\parindent{1em} Beyond reprojection accuracy, the intrinsic parameters estimated using RACE-FPP exhibited greater stability across all three datasets. The estimated focal lengths ($f_x$, $f_y$), principal point coordinates ($c_x$, $c_y$), and distortion coefficients showed only minor variations between the three independent characterisations, whereas noticeably larger fluctuations were observed for the OpenCV-based pipeline under degraded imaging conditions. The estimated camera intrinsic parameters for the three datasets are summarised in Table~\ref{tab:camera_intrinsics}, confirming that RACE-FPP yields more stable camera intrinsic parameters across the different imaging conditions.
\par \setlength\parindent{1em} All the aforementioned findings underline the link between detection accuracy and camera characterisation reliability. Accurate and structurally coherent keypoints not only minimise reprojection error but also ensure stable parameter estimation even when the imaging conditions deviate from ideal laboratory settings. In practical terms, this means that a single model trained on diverse image conditions can effectively synergise characterisation routines that traditionally rely on carefully controlled data acquisition.

\paragraph{Projector and stereo characterisation}
\hspace*{0.5em} Following camera characterisation, the performance of the proposed approach is further evaluated through projector and stereo characterisation. As described in the \textbf{Centre-based projector characterisation} subsection, the detected checkerboard corners are used to determine the centres of the white squares by averaging the coordinates of their four surrounding corners. The corresponding absolute phase values are then obtained at these sub-pixel centre locations using bilinear interpolation and mapped to projector coordinates. This centre-based strategy enables phase information to be sampled within the white-square regions, where stronger fringe modulation is observed, rather than directly at checkerboard corners. The projector parameters are subsequently estimated using OpenCV's \texttt{calibrateCamera} function. Stereo characterisation then determines the relative rotation and translation between the camera and projector, with the camera coordinate system used as the reference for subsequent 3D reconstruction. Table~\ref{tab:proj_stereo_rpe} summarises the root mean square reprojection errors obtained for projector and stereo characterisation using the same three image datasets evaluated previously.

\begin{table}[!t]
\centering
%\scriptsize
\footnotesize
\setlength{\tabcolsep}{11pt}
\renewcommand{\arraystretch}{1.75}
\caption{Reprojection error (RPE) in pixels (px) for projector and stereo characterisation using RACE-FPP and the OpenCV pipeline.}
\label{tab:proj_stereo_rpe}
\resizebox{\linewidth}{!}{%
\begin{tabular}{lcccc}
\toprule
\multirow{2}{*}{Dataset} 
& \multicolumn{2}{c}{RACE-FPP RPE} 
& \multicolumn{2}{c}{OpenCV RPE} \\
\cmidrule(lr){2-3} \cmidrule(lr){4-5}
& Projector & Stereo & Projector & Stereo \\
\midrule

Noise-free & 0.0512 & 0.1646 & 0.0357 & 0.1323 \\
Mixed images     & 0.3935 & 0.3778 & 0.8571 & 1.6647 \\
Noisy images     & 0.6305 & 0.1518 & 1.6849 & 3.1638 \\

\bottomrule
\end{tabular}
}
\end{table}

\par \setlength\parindent{1em} Under noise-free conditions, both pipelines achieve low projector and stereo characterisation reprojection errors, with OpenCV yielding slightly lower values of 0.0357 px and 0.1323 px, respectively, compared with 0.0512 px and 0.1646 px for the proposed approach. This behaviour is consistent with the camera characterisation results, where conventional sub-pixel corner localisation performs particularly well for clean, high-contrast checkerboard images. As the image quality deteriorates, however, a clear divergence between the two pipelines is observed. For the mixed dataset, the projector reprojection error obtained using the proposed approach is 0.3935 px compared with 0.8571 px for OpenCV, while the corresponding stereo errors are 0.3778 px and 1.6647 px, respectively. Under the noisy condition, the difference increases further as the proposed approach yields a projector error of 0.6305 px compared with 1.6849 px for OpenCV, while the stereo reprojection error remains at 0.1518 px compared with 3.1638 px for the conventional pipeline.

\begin{figure*}[!htbp]
    \begin{adjustbox}{minipage=\linewidth-4pt,rndframe={width=1pt,color=aa}{5pt}}
        \centering
        \includegraphics[width=0.9\linewidth]{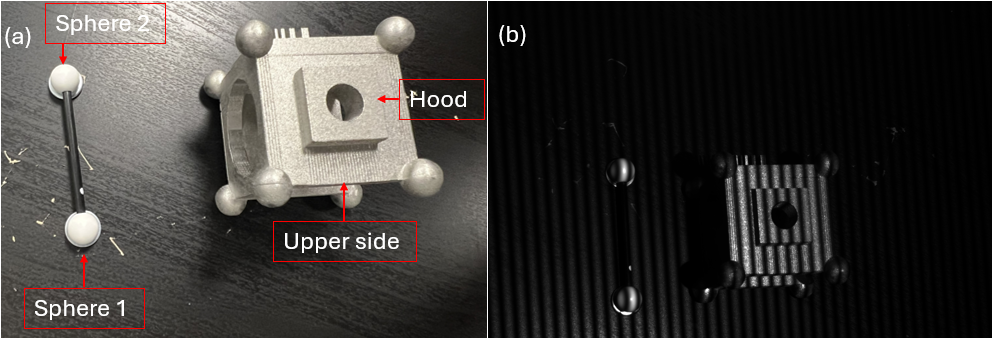}
        \caption{The measured artefacts: (a) raw image and (b) high-frequency fringes image. The dumbbell sphere is on the left, and the multi-feature artefact is on the right in each image.}
        \label{fig:artefacts}
    \end{adjustbox}
\end{figure*}

\begin{figure*}[h!]
    \begin{adjustbox}{minipage=\linewidth-4pt,rndframe={width=1pt,color=aa}{5pt}}    
        \centering
        \includegraphics[scale=0.52, trim=1.3cm 3.5cm 1.2cm 1.5cm, clip]{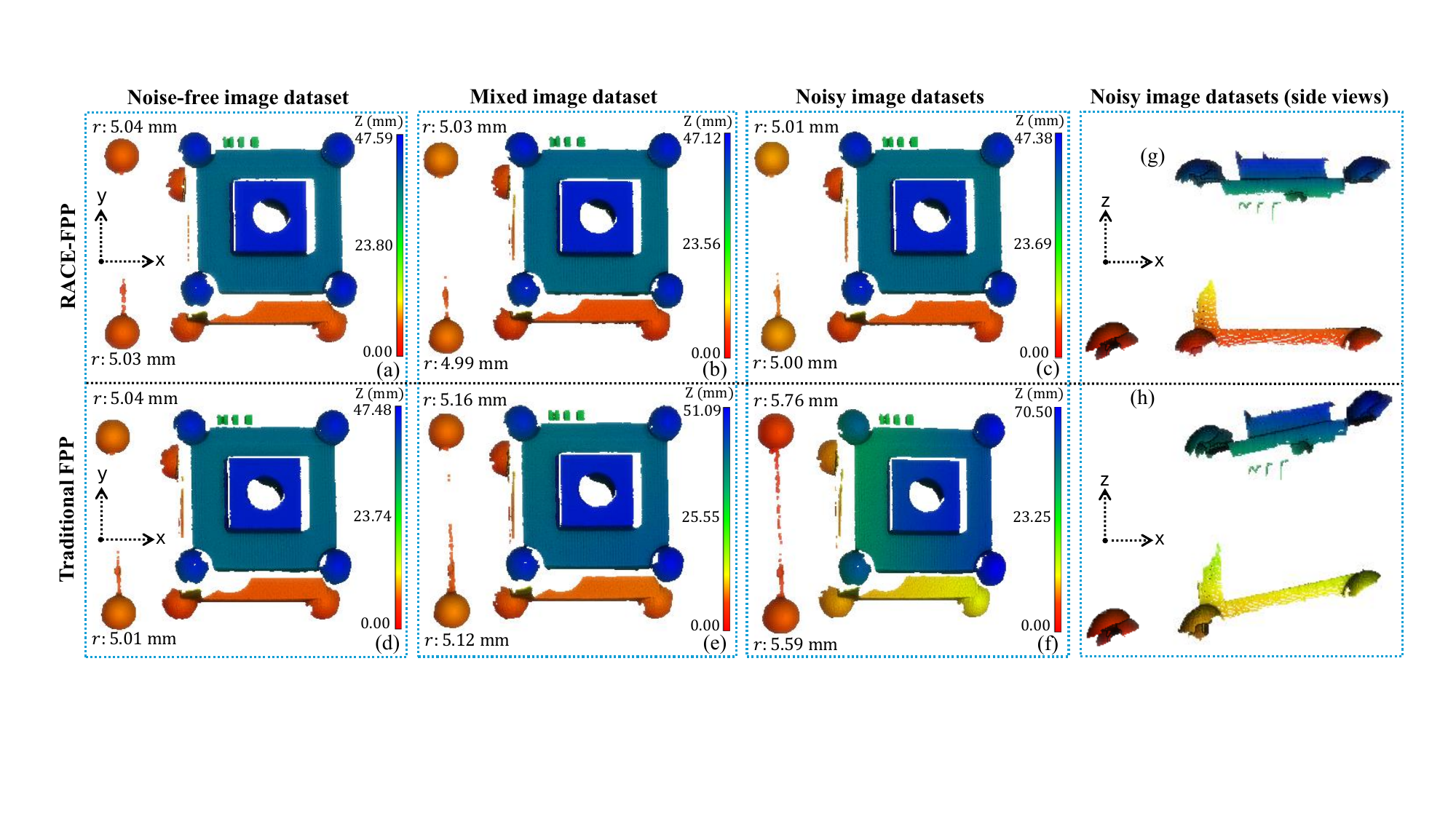}
        \caption{3D reconstructions obtained using the proposed and conventional characterisation pipelines: (a–c) RACE-FPP and (d–f) Traditional FPP for noise-free, mixed, and noisy image datasets, respectively; (g) and (h) side views of the corresponding noisy-condition reconstructions, highlighting differences in geometric distortion.}
        \label{fig:measurements}
    \end{adjustbox}
\end{figure*}

\begin{table*}[!htb]
\centering
\footnotesize
\setlength{\tabcolsep}{9pt}
\renewcommand{\arraystretch}{1.25}
\caption{Sphere radius and spacing distance (SD) measurements (mm) of the precision dumbbell sphere, with deviations from the corresponding CMM measurements.}
\label{tab:sd_sphere_dataset2}

\resizebox{\linewidth}{!}{%
\begin{tabular}{lcccccc}
\toprule
\textbf{Feature}& \textbf{Dataset} & \textbf{CMM} & \textbf{RACE-FPP} & \textbf{Deviation} & \textbf{Traditional FPP} & \textbf{Deviation} \\
\midrule

\multirow{3}{*}{SD} 
& Noise-free & \multirow{3}{*}{49.945} & 49.89 & -0.056 & 49.93 & -0.015 \\
& Mixed images  &     & 49.87 & -0.075 & 51.37 & +1.425 \\
& Noisy images  &     & 49.83 & -0.115 & 54.33 & +4.385 \\
\midrule

\multirow{3}{*}{Sphere 1 radius} 
& Noise-free & \multirow{3}{*}{4.997} & 5.03 & +0.033 & 5.01 & +0.013 \\
& Mixed images  &     & 4.99 & -0.007 & 5.12 & +0.123 \\
& Noisy images  &     & 5.00 & +0.003 & 5.59 & +0.593 \\
\midrule

\multirow{3}{*}{Sphere 2 radius} 
& Noise-free & \multirow{3}{*}{4.996} & 5.04 & +0.044 & 5.04 & +0.044 \\
& Mixed images  &    & 5.03 & +0.034 & 5.16 & +0.164 \\
& Noisy images  &    & 5.01 & +0.014 & 5.76 & +0.764 \\

\bottomrule
\end{tabular}
}
\end{table*}

\par \setlength\parindent{1em} The improved projector characterisation under degraded imaging conditions benefits from both the robustness of the detected checkerboard features and the centre-based projector correspondence strategy. Deriving each white-square centre from four surrounding detected corners reduces the influence of individual localisation deviations, while sampling the corresponding phase within the white-square region avoids direct reliance on the checkerboard intensity-transition regions, where weaker fringe modulation is observed. In contrast, the conventional pipeline relies directly on detected checkerboard corners and their corresponding phase values; consequently, degraded corner localisation and less reliable phase information can both affect the resulting projector correspondences.

\par \setlength\parindent{1em} The stereo results further demonstrate the stability of the estimated camera–projector geometry. Although OpenCV achieves a slightly lower stereo reprojection error under noise-free conditions, its error increases substantially for the mixed and noisy datasets, reaching 3.1638 px for the latter. In comparison, the proposed pipeline maintains sub-pixel stereo reprojection errors across all three conditions. In addition, the proposed detector successfully processes all characterisation images without dataset-specific adjustment of detection parameters, preserving the available characterisation poses under degraded imaging conditions. These results indicate that the proposed characterisation strategy maintains a more stable camera–projector relationship as image quality deteriorates, the dimensional consequences of which are examined through 3D reconstruction in the following subsection.

\begin{table*}[h!]
\centering
\footnotesize
\setlength{\tabcolsep}{9pt}
\renewcommand{\arraystretch}{1.25}
\caption{Average hood step-height measurements (mm) for the multi-feature artefact, with standard deviations (std) and deviations from the CMM reference.}
\label{tab:Hood step}

\resizebox{\linewidth}{!}{%
\begin{tabular}{lccccccc}
\toprule
\textbf{Dataset} & \textbf{CMM} &
\textbf{RACE-FPP} & \textbf{std} & \textbf{Deviation} &
\textbf{Traditional FPP} & \textbf{std} & \textbf{Deviation} \\
\midrule
Noise-free   & \multirow{3}{*}{4.917} & 4.91 & 0.04 & -0.007 & 4.90 & 0.05 & -0.017 \\
Mixed images &                        & 4.91 & 0.03 & -0.007 & 5.07 & 0.05 & +0.153 \\
Noisy images &                        & 4.92 & 0.04 & +0.003 & 4.97 & 0.05 & +0.053 \\
\bottomrule
\end{tabular}
}
\end{table*}

\paragraph{3D reconstruction and dimensional validation}
\hspace*{0.5em} The final validation of the proposed characterisation pipeline is performed through the 3D reconstruction of two artefacts: a precision dumbbell sphere, used as a standard reference artefact, and an additively manufactured component incorporating multiple geometric features (Figure~\ref{fig:artefacts}). For each image dataset -noise-free, mixed, and noisy- the characterisation parameters are estimated independently, resulting in six 3D reconstructions, as shown in Figure~\ref{fig:measurements}. The reconstructed point clouds have an average spatial resolution of approximately 50 $\mu m$.

\par \setlength\parindent{1em} The two artefacts provide complementary validation of the reconstruction performance. The precision dumbbell provides a controlled metrological reference through measurements of the sphere radii and centre-to-centre spacing distance (SD), enabling assessment of the reconstructed feature dimensions and relative spacing. The multi-feature artefact provides additional geometric validation through measurement of the step height between the planar surfaces of the Hood feature and Upper side, assessing the ability of the reconstruction to preserve the separation between surfaces at different depths. Visual inspection indicates that RACE-FPP produces geometrically consistent reconstructions across the different datasets, whereas distortions and misalignment become increasingly apparent in the Traditional FPP reconstructions under noisy conditions. In particular, the side view in Figure~\ref{fig:measurements}(h) highlights the geometric distortion in the Traditional FPP reconstruction obtained from the noisy dataset. For quantitative evaluation, the dumbbell spheres are fitted using a sphere-fitting algorithm to determine their radii and centre-to-centre spacing, while the step height of the multi-feature artefact is determined as the mean perpendicular distance between the hood-top point cloud and the plane fitted to the surrounding upper surface. These measurements are compared with the corresponding CMM reference values, with the resulting deviations reported in Tables~\ref{tab:sd_sphere_dataset2} and~\ref{tab:Hood step}.

\par \setlength\parindent{1em} Table~\ref{tab:sd_sphere_dataset2} presents the dimensional evaluation of the precision dumbbell sphere, a calibrated reference artefact with known geometry. Under noise-free conditions, both RACE-FPP and Traditional FPP produce comparable measurements with deviations below 0.05 mm, indicating that both characterisation pipelines achieve high reconstruction accuracy when corner localisation is reliable. As the image quality deteriorates, however, the behaviour of the two methods diverges significantly. RACE-FPP maintains stable reconstruction performance across all datasets, with sphere radius deviations remaining within approximately 0.04 mm and the SD deviation increasing only marginally to -0.115 mm under the noisy condition. In contrast, the Traditional FPP geometry exhibits a progressive increase in both local and global dimensional errors, with sphere radius deviations reaching up to 0.764 mm and the SD deviation increasing to 4.385 mm. The larger deviation observed in the SD measurement compared with the sphere radii suggests that the accumulated effects of corner localisation errors become increasingly pronounced when reconstructing the relative positions of spatially separated features. Although both measurements depend on the characterisation accuracy, the sphere radii are estimated from locally fitted surface points, whereas the SD additionally reflects the preservation of the overall reconstructed geometry. Consequently, inaccuracies in the camera, projector, and stereo characterisation are more readily manifested in the centre-to-centre distance than in the fitted sphere radii.
\par \setlength\parindent{1em} Table~\ref{tab:Hood step} presents the step-height measurements obtained from the multi-feature artefact. RACE-FPP results remain consistent across the three characterisation datasets, with measured heights of 4.91, 4.91, and 4.92 mm for the noise-free, mixed, and noisy conditions, respectively, compared with the CMM reference value of 4.917 mm. The corresponding deviations remain within 0.01 mm across all conditions, indicating that the reconstructed depth separation is maintained despite degradation of the characterisation images. In comparison, the Traditional FPP measurements show greater variation, with deviations ranging from $-0.02$ to $+0.15$ mm and the largest deviation occurring for the mixed-image dataset. These results provide additional evidence that the proposed characterisation approach maintains dimensional stability under the investigated image-degradation conditions.
\par \setlength\parindent{1em} The 3D reconstruction and dimensional evaluations demonstrate that both characterisation pipelines provide comparable performance under noise-free conditions, whereas their behaviour diverges substantially as image quality deteriorates. The proposed characterisation framework maintains geometrically consistent reconstructions and bounded dimensional deviations across both reference artefacts, while the conventional pipeline exhibits greater geometric distortion and substantially larger dimensional deviations under degraded conditions. These findings demonstrate that the improved stability of the proposed camera–projector characterisation is reflected in the dimensional fidelity of the reconstructed geometry.

\section*{Conclusion}
\noindent
This work developed a robust characterisation framework for FPP systems operating under non-ideal imaging conditions, particularly where regions of varying image quality coexist within the same frame due to object orientation, depth variation, and optical effects. The framework combines deep learning-based checkerboard corner localisation with a centre-based projector characterisation strategy, while retaining the conventional geometric camera–projector model. White-square centres are derived from the detected checkerboard geometry and used to sample phase information for establishing camera–projector correspondences. The approach therefore modifies the feature-localisation and projector-correspondence stages without replacing the underlying physical characterisation model.
\par \setlength\parindent{1em} The proposed framework demonstrated substantially greater robustness as image quality deteriorated. Although the conventional OpenCV pipeline achieved slightly lower reprojection errors for clean, high-contrast images, its performance degraded markedly under lens blur and noise, with several checkerboard images no longer successfully detected. In contrast, the proposed method successfully processed all characterisation images and maintained substantially lower reprojection errors under degraded conditions. For the noisy-image dataset, the camera characterisation reprojection error declined from 2.85 px with the conventional pipeline to 0.45 px, while the projector characterisation reprojection error decreased from 1.68 px to 0.63 px. Sub-pixel stereo reprojection errors were maintained across all tested imaging conditions, indicating greater stability of the estimated camera–projector geometry.
\par \setlength\parindent{1em} The practical significance of this robustness was confirmed through 3D reconstruction and CMM-referenced dimensional measurements. For a precision dumbbell artefact, the centre-to-centre spacing deviation under noisy conditions was -0.115 mm with the proposed framework, compared with 4.385 mm for the conventional pipeline. Measurements of a step height on a multi-feature artefact further demonstrated dimensional stability across the different characterisation conditions, with deviations from the CMM reference remaining within 0.01 mm for the proposed framework. These results demonstrate that the increased stability of the proposed characterisation framework extends beyond image-domain reprojection metrics to the dimensional fidelity of the reconstructed geometry.
\par \setlength\parindent{1em} Overall, the results show that combining data-driven feature localisation with centre-based projector correspondences can substantially improve the robustness of the complete FPP characterisation chain while preserving its geometric and physically interpretable formulation. This is particularly relevant to practical dimensional measurements using FPP, where spatially non-uniform blur, noise, and other image degradations cannot always be avoided.
\par \setlength\parindent{1em} The present study considered a single FPP configuration. Future work will therefore examine transferability across different camera resolutions, fields of view, optical geometries, and measurement systems. It will also investigate the influence of training-data composition on the balance between performance under ideal conditions and robustness to image degradation, with the aim of reducing the small accuracy gap observed for clean images without sacrificing performance under challenging acquisition conditions.

\begin{funding}
 United Kingdom Research and Innovation - Engineering \& Physical Sciences Research Council (EPSRC); Taraz Metrology Ltd., UK.   
\end{funding}

\begin{acknowledgement}
    \noindent
    Thanks to the EPSRC and Taraz Metrology Ltd. for funding this work. Thanks are also extended to Xiangjun Kong and Tibebe Yalew with the EPSRC project number EP/X010929/1. Also, thanks to Mohamed Isa from the Manufacturing Metrology Team (MMT) for the CMM measurements, Mark East from the Centre for Additive Manufacturing (CfAM), University of Nottingham, for 3D printing the multi-feature artefact, and Sean Lau and Qian Kai Loh from Taraz Metrology Ltd for the assistance with the CMM measurement of the dumbbell sphere and FPP system backend work.  
\end{acknowledgement}

\begin{authorcontributions}
    \noindent
    Osman Ali: Conceptualization, data collection, methodology, experiment, analysis, and original draft. Xiangjun Kong: Conceptualization, draft review and editing. Tibebe Yalew: Deep learning model proposal and draft review. Waiel Elmadih: Funding acquisition, industrial supervision, draft review. Samanta Piano: Funding acquisition, draft review and editing.
\end{authorcontributions}

%\begin{Data availability}
 %   Data from this work are available upon reasonable request.
%\end{Data availability}

\begin{conflictofinterest}
    \noindent
    The authors declare no conflict of interest. 
\end{conflictofinterest}

\bibliography{ref}
%\nocite{*}

\end{document}